\documentclass[leqno,tbtags]{sig-alternate}
\usepackage[logonly]{trace}
\usepackage{fixltx2e}
\usepackage{amsmath1}
\usepackage{amstext}
\usepackage{amssymb}
\usepackage{mathdots}

\usepackage{ifpdf}
\ifpdf
    \usepackage[pdf]{qtree}
    \usepackage[pdftex,matrix,arrow,curve,frame]{xy}
\else
    \usepackage{qtree}
    \usepackage[ps,dvips,matrix,arrow,curve,frame]{xy}
\fi
\entrymodifiers={+!!<0pt,\the\fontdimen22\textfont2>}

\makeatletter
\newcommand{\shrink}[1][\jot]{\relax \setlength\@tempdima{#1}%
  \ifmmode \expandafter\mathpalette\expandafter\math@shrink
  \else \expandafter\make@shrink \fi}
\newcommand{\make@shrink}[1]{\setbox\z@\hbox{\kern-\@tempdima\color@begingroup#1\color@endgroup\kern-\@tempdima}\fin@shrink}
\newcommand{\math@shrink}[2]{\setbox\z@\hbox{\kern-\@tempdima$\m@th#1{#2}$\kern-\@tempdima}\fin@shrink}
\newcommand{\fin@shrink}{%
  \@tempdimb\ht\z@ \advance\@tempdimb-\@tempdima \ht\z@\@tempdimb
  \@tempdimb\dp\z@ \advance\@tempdimb-\@tempdima \dp\z@\@tempdimb
  \box\z@}
\makeatother

\usepackage{hyphenat}
\usepackage{exemplar}
\usepackage{array1}
\usepackage{qtree1}
\usepackage{url}
\usepackage{varioref}
\usepackage{refrange}
\usepackage{calc}
\usepackage{prooftree}
\usepackage{mdwlist}

\usepackage{flushend}

\makeatletter \providecommand{\newblock}{\@undefined} \makeatother
\usepackage[longnamesfirst,sort&compress]{natbib1}
\bibsep=\smallskipamount
\renewcommand\bibsection{\section{\MakeUppercase{\refname}}}

\usepackage{soul}

\usepackage{semantics}
\renewcommand{\emp}{\emph}
\renewcommand{\denotation}[1]{\text{\sffamily#1}}
\DeclareDenotation{\Student}{student}
\DeclareDenotation{\Diligent}{diligent}
\DeclareDenotation{\Passed}{passed}
\DeclareDenotation{\Course}{course}
\DeclareDenotation{\Alice}{alice}
\DeclareDenotation{\CSnlp}{cs187}
\DeclareDenotation{\Liked}{liked}
\DeclareDenotation{\Left}{left}
\DeclareDenotation{\Representative}{repr}
\DeclareDenotation{\Of}{of}
\DeclareDenotation{\Company}{company}
\DeclareDenotation{\Animate}{animate}
\DeclareDenotation{\Thing}{Thing}
\DeclareDenotation{\Bool}{Bool}
\DeclareDenotation{\BoolNeg}{BoolNeg}
\DeclareDenotation{\BoolPos}{BoolPos}

\DeclareMathVersion{code}
\SetMathAlphabet{\mathnormal}{code}{OT1}{cmtt}{m}{n}
\SetMathAlphabet{\mathbf}{code}{OT1}{cmtt}{b}{n}
\SetMathAlphabet{\mathsf}{code}{OT1}{cmtt}{m}{n}
\SetMathAlphabet{\mathit}{code}{OT1}{cmtt}{m}{it}
\SetSymbolFont{operators}{code}{OT1}{cmtt}{m}{n}
\SetSymbolFont{letters}{code}{OT1}{cmtt}{m}{n}  

\newcommand{\Colon     }{.\,}

\newcommand{\Shift     }{\xi}
\newcommand{\Lam       }{\lambda}

\newcommand{\fwd}{\mathbin\prime}
\newcommand{\bwd}{\mathbin\backprime}
\newcommand{\Fwd}{\mathbin{\acute\rightarrow}}
\newcommand{\Bwd}{\mathbin{\grave\rightarrow}}
\makeatletter
\renewcommand{\fun}[2][]{\mathopen{\smash[t]{\lambda
      ^{\text{\setbox\z@=\hbox{$\!#1\!$}%
              \ifdim\wd\z@>\z@\unhbox\z@\else\hb@xt@\z@{\hss$#1$\hss}\fi}}%
    }\mathord{#2}.\,}}
\makeatother

\newcommand{\evalsto}{\mathrel\vartriangleright}
\newcommand{\toD}{\mathbin{\smash[t]{\stackrel{\Delta}%
    {\text{\raisebox{0pt}[.8\height-.2\depth][.8\depth-.2\height]%
                         {$\rightarrow$}}}}}}
\newcommand{\toMaybeD}{\mathbin{\smash[t]{\stackrel{\smash[b](\Delta\smash[b])}%
    {\text{\raisebox{0pt}[.8\height-.2\depth][.8\depth-.2\height]%
                         {$\rightarrow$}}}}}}

\makeatletter
\def\@sect#1#2#3#4#5#6[#7]#8{%
    \ifnum #2>\c@secnumdepth
        \let\@svsec\@empty
    \else
        \refstepcounter{#1}%
        \edef\@svsec{%
            \begingroup
			\ifnum#2>2 \noexpand#6 \fi
                \csname the#1\endcsname
            \endgroup
            \ifnum #2=1\relax .\fi
            \hskip 1em
        }%
    \fi
    \@tempskipa #5\relax
    \ifdim \@tempskipa>\z@
        \begingroup
            #6\relax
            \@hangfrom{\hskip #3\relax\@svsec}%
            \begingroup
                \interlinepenalty \@M
                \if@uchead
                    \uppercase{#8}%
                \else
                    #8%
                \fi
                \par
            \endgroup
        \endgroup
        \csname #1mark\endcsname{#7}%
\addcontentsline{toc}{#1}{%
            \ifnum #2>\c@secnumdepth \else
                \protect\numberline{\csname the#1\endcsname}%
            \fi
            #7%
        }%
    \else
        \def\@svsechd{%
            #6%
            \hskip #3\relax
            \@svsec
            \if@uchead
                \uppercase{#8}%
            \else
                #8%
            \fi
            \csname #1mark\endcsname{#7}%
            \addcontentsline{toc}{#1}{%
                \ifnum #2>\c@secnumdepth \else
                    \protect\numberline{\csname the#1\endcsname}%
                \fi
                #7%
            }%
        }%
    \fi
    \@xsect{#5}
    \par
}
\def\section{%
    \@startsection{section}{1}{\z@}{10\p@ \@plus 4\p@ \@minus 2\p@}
    {4\p@}{\baselineskip 14pt\secfnt\@ucheadtrue}%
}
\def\subsection{%
    \@startsection{subsection}{2}{\z@}{8\p@ \@plus 2\p@ \@minus \p@}
    {4\p@}{\secfnt}%
}
\def\subsubsection{%
    \@startsection{subsubsection}{3}{\z@}{8\p@ \@plus 2\p@ \@minus \p@}%
    {4\p@}{\subsecfnt}%
}
\let\@@category\@category
\def\@category{\noindent\@@category}
\makeatother

\begin{document}

\conferenceinfo{Continuation Workshop}{2004 Venice, Italy}
\title{Delimited continuations in natural language}
\subtitle{Quantification and polarity sensitivity}
\numberofauthors{1}
\author{
\alignauthor Chung-chieh Shan\\[\jot]
    \affaddr{Harvard University}\\
    \affaddr{33 Oxford Street}\\
    \affaddr{Cambridge, MA 02138 USA}\\
    \email{ccshan@post.harvard.edu}}
\maketitle

\begin{abstract}
\noindent
Making a linguistic theory is like making a programming language: one typically
devises a type system to delineate the acceptable utterances and a denotational
semantics to explain observations on their behavior.  Via this connection, the
programming language concept of delimited continuations can help analyze
natural language phenomena such as quantification and polarity sensitivity.
Using a logical metalanguage whose syntax includes control operators and whose
semantics involves evaluation order, these analyses can be expressed in direct
style rather than continuation\hyp passing style, and these phenomena can be
thought of as computational side effects.
\end{abstract}


\noindent{Languages, Theory}

\keywords\noindent{Delimited continuations, control effects, natural language
semantics, quantification, polarity sensitivity}

\section{Introduction}
\label{s:introduction}

This paper is about computational linguistics, in the sense of applying
insights from computer science to linguistics.  Linguistics strives to
scientifically explain empirical observations of natural language.
Semantics, in particular, is concerned with phenomena such as the
following.  In~\eqref{e:entailment} below, some sentences to the left
\emp{entail} their counterparts to the right, but others do not.
\example{\label{e:entailment}%
    \subexamples[l@{\;}C@{\;}X]{
        Every student passed
        & \vdash
        & Every diligent student passed \\
        No student passed
        & \vdash
        & No diligent student passed \\
        A student passed
        & \nvdash
        & A diligent student passed \\
        Most students passed
        & \nvdash
        & Most diligent students passed}}
The sentence in~\eqref{e:ambiguity} is \emp{ambiguous} between at least
two readings.  On one reading, the speaker must decline to run any spot
that fails to substantiate any claims whatsoever.  On another reading, there
exist certain claims (anti-war ones, say) such that the speaker must decline to
run any spot that fails to substantiate them.
\example{\label{e:ambiguity}%
    We must decline to run any spot that fails to substantiate certain claims.%
    \footnotemark}
\footnotetext{%
    This sentence is part of a statement made by the cable television
    company Comcast after its CNN channel rejected an anti-war commercial hours
    before it was scheduled to air on January 28, 2003.}%
Finally, among the four sentences in~\eqref{e:acceptability},
only~\eqref{e:acceptability-acceptable} is \emp{acceptable}.  That is, only
it can be used in idealized conversation.  The unacceptability of the rest
is notated with asterisks.
\example{\label{e:acceptability}%
    \subexamples{
    \subexample{\label{e:acceptability-acceptable}%
                   No    student  liked any course.}
    \subexample{\<*Every student  liked any course.}
    \subexample{\<*A     student  liked any course.}
    \subexample{\<*Most  students liked any course.}}}

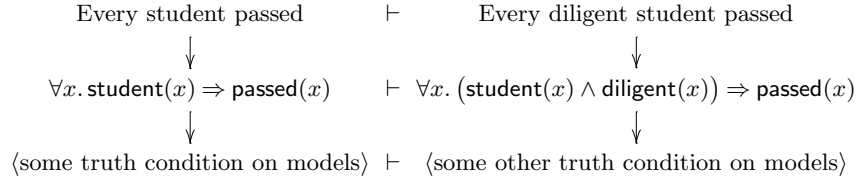
\begin{figure*}[t]
\smallskip
\centerline{\shrink{$\xymatrix @C=0pt @R=3ex{
    \strut\text{Every student passed} \ar[d]
    &\vdash&
    \strut\text{Every diligent student passed} \ar[d]
\\
    \strut\Forall{x} \Student(x) \limplies \Passed(x) \ar[d]
    &\vdash&
    \strut\Forall{x} \bigl( \Student(x) \land \Diligent(x) \bigr) \limplies \Passed(x) \ar[d]
\\
    \strut\langle\text{some truth condition on models}\rangle
    &\vdash&
    \strut\langle\text{some other truth condition on models}\rangle
    }$}}
\caption{The translation\slash denotation approach to natural language semantics}
\label{fig:translation}
\end{figure*}

The linguistic entailments and non\hyp entailments in~\eqref{e:entailment} are
facts about English, in that only a speaker of English can make these judgments.
Nevertheless, they presumably have to do with corresponding logical entailments
and non\hyp entailments: both the English speaker who judges that \phrase{Every
student passed} entails \phrase{Every diligent student passed} and the Mandarin
speaker who judges that \phrase{Meige xuesheng dou jige-le} entails
\phrase{Meige yonggong-de xuesheng dou jige-le} rely on knowing that,
if every student passed, then every diligent student passed.
Thus the typical linguistic theory specifies a semantics for natural
language by translating declarative sentences into logical statements
with truth conditions.  The linguistic entailments
in~\eqref{e:entailment} hold, goes the theory, because the
meanings---truth conditions---of the two sentences are such that any model
that verifies the former also verifies the latter.
Much work in natural language semantics aims in this way, as depicted in
Figure~\vref{fig:translation}, to explain the horizontal by positing the
vertical.  This approach is
reminiscent of programming language research where an ill-understood
language (perhaps one with a complicating feature like exceptions) is
studied by translation into a simpler language (without exceptions) that is
better understood.

The translation target posited in natural language semantics is often some
combination of the $\lambda$-calculus and predicate logic.  For example, the
verb \phrase{passed} might be translated as $\fun{x}\Passed(x)$.  This paper
argues by example that the translation target should be a logical metalanguage
with first-class delimited continuations.  The examples are two natural
language phenomena: \emp{quantification} by words like \phrase{every} and
\phrase{most} in~\eqref{e:entailment}, and \emp{polarity sensitivity} on the
part of words like \phrase{any} in~\eqref{e:acceptability}.

Quantification was first analyzed explicitly using continuations by
\citet{barker-continuations}.  Building on that insight, this paper makes the
following two contributions.  First, I analyze natural language in direct style
rather than in continuation\hyp passing style.  In other words, the logical
metalanguage used here is one that includes control operators for delimited
continuations, rather than a pure $\lambda$-calculus in which denotations need
to handle continuations explicitly.  Natural language is thus endowed with an
operational semantics from computer science that is richer than just
$\beta\eta$-reduction.

Second, I propose a new analysis of polarity sensitivity that improves upon
prior theories in explaining why \phrase{No student liked any course}
is acceptable but *\phrase{Any student liked no course} is not.  This
analysis crucially relies on the notion of evaluation order from
programming languages, thus elucidating the role of control effects
in natural language and supporting the broader claim that linguistic
phenomena can be fruitfully thought of as computational side effects.

The rest of this paper is organized as follows.  In~\S\ref{s:formalism},
I introduce a simple grammatical formalism.  In~\S\ref{s:quantification},
I describe the linguistic phenomenon of quantification and show a straw man
analysis that deals with some cases but not others.  I then introduce
a programming language with delimited continuations and use it to improve the straw
man analysis: quantification in non-subject position is
treated in~\S\ref{s:delimited-continuations}, and inverse scope is covered
in~\S\ref{s:hierarchy}.  In~\S\ref{s:polarity}, I turn to the linguistic
phenomenon of polarity sensitivity and show how a computationally motivated
notion of evaluation order improves upon previous analyses.
In~\S\ref{s:beyond}, I place these examples in a broader context and conclude.

\section{A grammatical formalism}
\label{s:formalism}

In this section, I introduce a simple grammatical formalism for use in the
rest of the paper.  It is a notational variant of categorial grammar (as
introduced by \citet[chapter~4]{carpenter-type-logical}, for instance).

The verb \phrase{like} usually requires an object to its right and
a subject to its left.
\example{\subexamples{
\subexample{\label{e:alice-liked-cs187}Alice liked CS187.}
\subexample{\label{e:alice-liked}\<*Alice liked.}
\subexample{\label{e:alice-liked-bob-cs187}\<*Alice liked Bob CS187.}}}
Intuitively, \phrase{like} is a function that takes two arguments, and the sentences
(\refrange{e:alice-liked}{e:alice-liked-bob-cs187}) are unacceptable
due to type mismatch.  We can model this formally by assigning types to the
denotations of \phrase{Alice}, \phrase{CS187}, and \phrase{liked}, which we
take to be atomic expressions.
\begin{alignat}{2}
    \denote{\text{Alice}} &= \Alice &&: \Thing \\
    \denote{\text{CS187}} &= \CSnlp &&: \Thing \\
    \denote{\text{liked}} &= \Liked &&: \Thing \toF \Thing \toF \Bool
\end{alignat}
Here $\Thing$ is the type of individual objects, and $\Bool$ is the type of
truth values or propositions.  Following (justifiable) standard practice in
linguistics, we let $\Liked$ take its object as the first argument and its
subject as the second argument.  For example,
in~\eqref{e:alice-liked-cs187}, the first argument to~$\Liked$ is $\CSnlp$,
and the second argument is~$\Alice$.

As \eqref{e:alice-liked-cs187} shows, there are two ways to combine
expressions.  A function can take its
argument either to its right (combining \phrase{liked} with \phrase{CS187})
or to its left (combining \phrase{Alice} with \phrase{liked CS187}).  We
denote these two cases with two infix operators: ``$\fwd$'' for forward
combination and ``$\bwd$'' for backward combination.  (The tick marks
depict the direction in which a function ``leans on'' an argument.)
\begin{alignat}{3}
\label{e:fwd-pre}
    f\fwd x &= f(x) &&: \beta &\qquad&\text{where $f:\alpha\toF\beta$,\enspace $x:\alpha$}\\
\label{e:bwd-pre}
    x\bwd f &= f(x) &&: \beta &\qquad&\text{where $f:\alpha\toF\beta$,\enspace $x:\alpha$}
\end{alignat}
We can now derive the sentence~\eqref{e:alice-liked-cs187}---that is, prove
it to have type~$\Bool$.  The derivation can be written as
a tree~\eqref{e:alc-tree} or a term~\eqref{e:alc-term}.
\begin{gather}
    \label{e:alc-tree}
    \leaf{Alice}
    \leaf{liked}
    \leaf{CS187}
    \emptybranch{2}
    \emptybranch{2}
    \qobitree
\\
    \label{e:alc-term}
    \denote{\text{Alice}} \bwd \left(\denote{\text{liked}} \fwd \denote{\text{CS187}}\right)
    = \Liked\;\CSnlp\;\Alice : \Bool
\end{gather}
By convention, the infix operators $\fwd$ and $\bwd$ associate to the
right, so parentheses such as those in~\eqref{e:alc-term}
above are optional.

Unfortunately, the system set up so far derives not only the acceptable
sentence~\eqref{e:alice-liked-cs187} but also the unacceptable
sentence~\eqref{e:alice-cs187-liked}, with the same meaning.
\example{\label{e:alice-cs187-liked}%
    \<*Alice CS187 liked.}
The reason the system derives~\eqref{e:alice-cs187-liked} is that the
direction of function application is unconstrained: in the derivation
below, \phrase{liked} takes its first (object) argument to the left, which
is usually disallowed in English.
\begin{gather}
    \label{e:acl-tree}
    \leaf{Alice}
    \leaf{CS187}
    \leaf{liked}
    \emptybranch{2}
    \emptybranch{2}
    \qobitree
\displaybreak[0]
\\
    \label{e:acl-term}
    \denote{\text{Alice}} \bwd \denote{\text{CS187}} \bwd \denote{\text{liked}}
    = \Liked\;\CSnlp\;\Alice : \Bool
\end{gather}
To rule out this derivation of~\eqref{e:alice-cs187-liked} in our type
system, we split the function type constructor ``$\toF$'' into two type
constructors ``$\Fwd$'' and~``$\Bwd$'', one for each direction of
application.  Using these new type constructors, we change the denotation
of \phrase{liked} to specify that its first argument is to its right and
its second argument is to its left.
\begin{equation}
    \denote{\text{liked}} = \Liked : \Thing \Fwd \Thing \Bwd \Bool
\end{equation}
We also revise the combination rules \eqref{e:fwd-pre}
and~\eqref{e:bwd-pre} to require different function type constructors.
\begin{alignat}{2}
\label{e:fwd}
    f\fwd x &: \beta &\qquad&\text{where $f:\alpha\Fwd\beta$,\enspace $x:\alpha$}\\
\label{e:bwd}
    x\bwd f &: \beta &\qquad&\text{where $f:\alpha\Bwd\beta$,\enspace $x:\alpha$}
\end{alignat}
The system now rejects~\eqref{e:alice-cs187-liked} while continuing to
accept~\eqref{e:alice-liked-cs187}, as desired.

\section{Quantification}
\label{s:quantification}

The linguistic phenomenon of quantification is illustrated by the following
sentences.
\example{
\subexamples{
\subexample{\label{e:eslc}Every student liked CS187.}
\subexample{\label{e:sslec}Some student liked every course.}
\subexample{\label{e:atbtm}Alice consulted Bob before most meetings.}}}
As with the previously encountered sentences, the natural language
semanticist wants to translate English into logical formulas that
account for entailment and other properties.  More precisely, the problem
is to posit translation rules that map these sentences thus.  For instance,
we would like to map~\eqref{e:eslc} to a formula like
\begin{equation}
\label{e:eslc-formula}
    \Forall{x} \Student(x) \limplies x\bwd\Liked\fwd\CSnlp : \Bool,
\end{equation}
where the constants
\begin{align}
    \forall  &:(\Thing\toF\Bool)\toF\Bool, &
    \limplies&:\Bool\toF\Bool\toF\Bool \hspace*{-.5em}
\end{align}
are drawn from the (higher-order) abstract syntax of predicate logic.  To this
end, what should the subject noun phrase \phrase{every student} denote?  Unlike
with \phrase{Alice}, there is nothing of type~$\Thing$ that the quantificational
noun phrase \phrase{every student} can denote and still allow the desired
translation~\eqref{e:eslc-formula} to be generated.  At the same time, we would
like to retain the denotation that we previously computed for the verb phrase
\phrase{liked CS187}, namely $\Liked\fwd\CSnlp$.  Taking
these considerations into account, one way to translate~\eqref{e:eslc}
to~\eqref{e:eslc-formula} is for the determiner \phrase{every} to denote
\begin{equation}
\label{e:every-term}
\begin{split}
    \denote{\text{every}}
    & = \fun[\fwd]{r} \fun[\fwd]{s} \Forall{x} r(x) \limplies x\bwd s \\
    & \eqish: (\Thing\toF\Bool)\Fwd(\Thing\Bwd\Bool)\Fwd\Bool.
\end{split}
\end{equation}
Here the \emp{restrictor}~$r$ and the \emp{scope}~$s$ are $\lambda$-bound
variables intended to receive, respectively, the denotations of the noun
\phrase{student} (of type $\Thing\toF\Bool$) and the verb phrase
\phrase{liked CS187} (of type $\Thing\Bwd\Bool$).  (More precisely, $r$
and~$s$ are $\lambda^{\!\fwd}$-bound variables; the tick mark again signifies
the direction of function application.)  In
a non-quantificational sentence like~\eqref{e:alice-liked-cs187}, the verb
phrase takes the subject as its argument; by contrast, in the
quantificational sentence~\eqref{e:eslc}, the subject takes the verb phrase
as its argument.

Extended with the lexical entry~\eqref{e:every-term} for \phrase{every},
and assuming that \phrase{student} denotes
\begin{equation}
    \denote{\text{student}} = \Student : \Thing\toF\Bool,
\end{equation}
the grammar can derive the sentence~\eqref{e:eslc}.
\begin{gather}
    \label{e:eslc-tree}
    \leaf{every}
    \leaf{student}
    \emptybranch{2}
    \leaf{liked}
    \leaf{CS187}
    \emptybranch{2}
    \emptybranch{2}
    \qobitree
\\
    \label{e:eslc-term}
    \left(\denote{\text{every}} \fwd \denote{\text{student}}\right) \fwd
          \denote{\text{liked}} \fwd \denote{\text{CS187}}
    = \eqref{e:eslc-formula}
\end{gather}
The existential determiner \phrase{some} can be analyzed similarly:
let \phrase{some} denote
\begin{equation}
\label{e:some-term}
\begin{split}
    \denote{\text{some}}
    &= \fun[\fwd]{r} \fun[\fwd]{s} \Exists{x} r(x) \land x\bwd s \\
    &\eqish: (\Thing\toF\Bool)\Fwd(\Thing\Bwd\Bool)\Fwd\Bool
\end{split}
\end{equation}
to derive the sentence \phrase{Some student liked CS187}.
\begin{gather}
    \label{e:sslc-tree}
    \leaf{some}
    \leaf{student}
    \emptybranch{2}
    \leaf{liked}
    \leaf{CS187}
    \emptybranch{2}
    \emptybranch{2}
    \qobitree
\displaybreak[0]
\\
    \label{e:sslc-term}
    \begin{split}
    \displayleft{\left(\denote{\text{some}} \fwd \denote{\text{student}}\right) \fwd
        \denote{\text{liked}} \fwd \denote{\text{CS187}}} \\
    \qquad&= \Exists{x} \Student(x) \land x\bwd\Liked\fwd\CSnlp : \Bool
    \end{split}
\end{gather}

To summarize, we treat determiners
like \phrase{every} and \phrase{some} as functions of two arguments: the
restrictor and the scope of a quantifier, both functions from~$\Thing$
to~$\Bool$.  Such higher-order functions are a popular analysis of natural
language determiners, and have been known to semanticists since
\citet{montague-proper} as \emp{generalized quantifiers}.  However, the
simplistic account presented above only handles quantificational noun
phrases in subject position, as in~\eqref{e:eslc} but not \eqref{e:sslec}
or~\eqref{e:atbtm}.  For example, in~\eqref{e:sslec}, neither forward nor
backward combination can apply to join the verb \phrase{liked}, of type
$\Thing\Fwd\Thing\Bwd\Bool$, to its object \phrase{every course}, of type
$(\Thing\Bwd\Bool)\Fwd\Bool$.  Yet, empirically speaking, the
sentence~\eqref{e:sslec} is not only acceptable but in fact ambiguous
between two available readings.  This problem has prompted a great variety
of supplementary proposals in the linguistics literature \citep[][inter
alia]{barwise-generalized,may-logical,hendriks-studied}.  The next
section presents a solution using \emp{delimited continuations}.

\section{Delimited continuations}
\label{s:delimited-continuations}

First-class continuations represent ``the entire (default) future for the
computation'' \citep{kelsey-r5rs-short}.  Refining this concept,
\citet{felleisen-theory} introduced \emph{delimited} continuations, which
encapsulate only a prefix of that future.  This paper uses \emp{shift} and
\emp{reset} \citep{danvy-functional,danvy-abstracting}, a popular choice of
control operators for delimited continuations.

To review briefly, the shift operator (notated~$\Shift$) captures the current
context of computation, removing it and making it available to the program as
a function.  For example, when evaluating the term
\begin{equation}
\label{e:shift-example}
    10 \times (\Shift f.\, 1 + f(2)),
\end{equation}
the variable~$f$ is bound to the function that multiplies every number
by~$10$.  Thus the above expression evaluates to~$21$ via the following
sequence of reductions.  (The reduced subexpression at each step is underlined.)
\begin{align}
    \rlap{$[10 \times (\text{\ul{$\Shift f.\, 1 + f(2)$}})]$}\enspace \\
    &\evalsto [1 + \text{\ul{$(\Lam v\Colon [10 \times v])(2)$}}] \notag\\
    &\evalsto [1 + [\text{\ul{$10 \times 2$}}]]
     \evalsto [1 + \text{\ul{$[20]$}}]
     \evalsto [\text{\ul{$1 + 20$}}]
     \evalsto \text{\ul{$[21]$}}
     \evalsto 21 \notag
\end{align}
Term reductions are performed deterministically in applicative order:
call-by-value and left-to-right.

The reset operator (notated with square brackets $[\enspace]$)
delineates how far shift can reach: shift captures the current
context of computation up to the closest dynamically enclosing reset.
Hence ``$3\;\times$'' below is out of reach.
\begin{align}
\label{e:reset-example}
    \rlap{$\bigl[3 \times [10 \times (\text{\ul{$\Shift f.\, 1 + f(2)$}})]\bigr]$}\enspace \\
    &\evalsto \bigl[3 \times [1 + \text{\ul{$(\Lam v\Colon [10 \times v])(2)$}}]\bigr] \notag\\
    &\evalsto \bigl[3 \times [1 + [\text{\ul{$10 \times 2$}}]]\bigr]
     \evalsto \bigl[3 \times [1 + \text{\ul{$[20]$}}]\bigr]
     \evalsto\dotsb
     \evalsto 63 \notag
\end{align}

Shift and reset come with an operational semantics (illustrated by the
reductions above) as well as a denotational semantics (via the CPS
transform).  That both kinds of semantics are
available is important to linguistics, because
the meanings of natural language expressions (studied in
semantics) need to be related to how humans process them (studied
in psycholinguistics).

Quantificational expressions in natural language can be thought of as phrases
that manipulate their context.  In a sentence like
\phrase{Alice liked CS187} \eqref{e:alice-liked-cs187}, the context of
\phrase{CS187} is the function mapping each thing~$x$ to the proposition that
Alice liked~$x$.  Compared to the proper noun \phrase{CS187}, what is special
about a quantificational expression like \phrase{every course} is that it
captures its surrounding context when used.
\example{\label{e:alec}Alice liked [every course].}
Thus, loosely speaking, the meaning of the sentence~\eqref{e:alec} no
longer has the overall shape $\Alice\bwd\Liked\fwd\dotsb$ once the occurrence
of \phrase{every course} is considered, much as the meaning of the
program~\eqref{e:shift-example} no longer has the overall shape $10
\times \dotsb$ once the shift expression is evaluated.  Let us add
shift and reset to the target language of our translation from English.  We can
then translate \phrase{every course} as
\begin{equation}
\label{e:every-course}
    \denote{\text{every course}}
    = \Shift s\Colon \Forall{x} \Course(x) \limplies s(x)
    : \Thing_{\Bool}^{\Bool}.
\end{equation}
The type notation \smash[b]{$\alpha_\gamma^\delta$} here indicates an~$\alpha$
with a control effect; the CPS transform maps it to
$(\alpha\toF\gamma)\toF\delta$.  The denotation of \phrase{every course} behaves
locally as a $\Thing$, but requires the current context to have the answer
type~$\Bool$ and maintains that answer type.

To see the new denotation~\eqref{e:every-course} in action, let us derive the
sentence~\eqref{e:alec}.  The type of \phrase{every course}
is~$\Thing_\Bool^\Bool$, similar to the type~$\Thing$ of~\phrase{CS187}, so the
derivation of~\eqref{e:alec} is analogous
to~(\refrange{e:alc-tree}{e:alc-term}).
\begin{gather}
    \label{e:alec-tree}
    \leaf{Alice}
    \leaf{liked}
    \leaf{every course}
    \emptybranch{2}
    \emptybranch{2}
    \qobitree
\displaybreak[0]\\
\label{e:alec-term}
\begin{split}
    \rlap{$\bigl[\denote{\text{Alice}} \bwd
                 \denote{\text{liked}} \fwd
                 \denote{\text{every course}}\bigr]$}\enspace\\
    &= [\Alice\bwd \Liked\fwd
      \text{\ul{$\Shift s\Colon \Forall{x} \Course(x) \limplies s(x)$}} ] \hspace*{-1em} \\
    &\evalsto [\Forall{x} \Course(x) \limplies \text{\ul{$(\fun{v}
        [\Alice\bwd\Liked\fwd v])(x)$}}]
      \evalsto\dotsb\\
    &\evalsto \Forall{x} \Course(x) \limplies \Alice\bwd\Liked\fwd x : \Bool
\end{split}
\end{gather}

Like the straw man analysis in~\S\ref{s:quantification}, the denotation
in~\eqref{e:every-course} generalizes to determiners other than
\phrase{every}: we can abstract the noun \phrase{course} out of
\phrase{every course}, and deal with \phrase{some student} similarly.
\begin{align}
\label{e:every}
    \denote{\text{every}}
        &= \fun[\fwd]{r} \Shift s\Colon \Forall{x} r(x) \limplies s(x),\\
\label{e:some}
    \denote{\text{some}}
        &= \fun[\fwd]{r} \Shift s\Colon \Exists{x} r(x) \land     s(x) \\
        &\eqish: \smash[t]{(\Thing\toF\Bool)\Fwd\Thing_{\Bool}^{\Bool}} \notag
\end{align}
(We require here that the restrictor~$r$ have the type $\Thing\toF\Bool$, not
a type of the form $\smash{\Thing\toF\Bool_\gamma^\delta}$, so $r$ cannot incur
control effects when applied to~$x$.  Any control effect in the restrictor,
such as induced by the quantificational noun phrase \phrase{a company} in the
sentence \phrase{Every representative of a company left}, must be contained
within reset.)

More importantly, unlike the straw man analysis, the new analysis works
uniformly for quantificational expressions in subject, object, and other
positions, such as in~(\refrange{e:eslc}{e:atbtm}).  Intuitively, this is
because shift captures the context of an expression no matter how deeply it is
embedded in the sentence.%
\footnote{No matter how deep, that is, up to the closest dynamically enclosing
    reset.  Control delimiters correspond to \emph{islands} in natural language
    \citep{barker-continuations}.}
By adding control operators for delimited
continuations to our logical metalanguage, we arrive at an analysis of
quantification with greater empirical coverage.%

\begin{figure*}[t]
\newcommand{\just}{\justifies}
\abovedisplayskip     =0pt
\abovedisplayshortskip=0pt
\begin{align*}
    \tag*{Directions} \Delta
        &::= \fwd \mid \bwd \\
    \tag*{Types} \alpha,\beta,\gamma,\delta
        &::= \Thing \mid \Bool \mid \alpha\toF\beta
        \mid \smash[t]{\alpha\toD\beta_\gamma^\delta} \\
    \tag*{Antecedents} \Gamma
        &::= x_1:\alpha_1,\dotsc,x_n:\alpha_n \\
    \tag*{Terms} E,F
        &::= c \mid x \mid \fun{x}E \mid \fun[\Delta]{x}E
        \mid FE \mid F\fwd E \mid E\bwd F \mid [E] \mid \Shift f\Colon E
\end{align*}

\abovedisplayskip     =\smallskipamount
\abovedisplayshortskip=\smallskipamount
\renewcommand{\just}{\justifies\strut}
\noindent Constants $c:\alpha$
\begin{gather*}
    \begin{prooftree}
        \just \forall : (\Thing\toF\Bool)\toF\Bool
    \end{prooftree}
    \qquad
    \begin{prooftree}
        \just \exists : (\Thing\toF\Bool)\toF\Bool
    \end{prooftree}
\\
    \begin{prooftree}
        \just \limplies : \Bool\toF\Bool\toF\Bool
    \end{prooftree}
    \qquad
    \begin{prooftree}
        \just \land : \Bool\toF\Bool\toF\Bool
    \end{prooftree}
\\
    \begin{prooftree}
        \just \Student : \Thing\toF\Bool
    \end{prooftree}
    \qquad
    \begin{prooftree}
        \just \Liked : \Thing\Fwd(\Thing\Bwd\Bool_\delta^\delta)_\gamma^\gamma
    \end{prooftree}
    \qquad
    \dotsm
\end{gather*}

\renewcommand{\just}{\strut\justifies\strut}
\noindent Pure expressions $\Gamma \vdash E:\alpha$
\begin{gather*}
    \begin{prooftree}
        c:\alpha
        \just \Gamma \vdash c:\alpha
        \using \text{Const}
    \end{prooftree}
    \qquad
    \begin{prooftree}
        \just \Gamma,x:\alpha \vdash x:\alpha
        \using \text{Var}
    \end{prooftree}
    \qquad
    \begin{prooftree}
        \Gamma \vdash E:\alpha_\alpha^\beta
        \just \Gamma \vdash [E]:\beta
        \using \text{Reset}
    \end{prooftree}
\\
    \begin{prooftree}
        \Gamma, x:\alpha \vdash E:\beta
        \just \Gamma \vdash \fun{x}E:\alpha\toF\beta
        \using \toF\text{I}
    \end{prooftree}
    \qquad
    \begin{prooftree}
        \Gamma, x:\alpha \vdash E:\beta_\gamma^\delta
        \just \Gamma \vdash \fun[\Delta]{x}E:\alpha\toD\beta_\gamma^\delta
        \using \toD\text{I}
    \end{prooftree}
    \qquad
    \begin{prooftree}
        \Gamma \vdash F:\alpha\toF\beta
        \quad \Gamma \vdash E:\alpha
        \just \Gamma \vdash FE:\beta
        \using \toF\text{E}
    \end{prooftree}
\end{gather*}

\belowdisplayskip     =0pt
\belowdisplayshortskip=0pt
\noindent Impure expressions $\Gamma \vdash E:\alpha_\gamma^\delta$
\begin{gather*}
    \begin{prooftree}
        \Gamma \vdash E:\alpha
        \just \Gamma \vdash E:\alpha_\gamma^\gamma
        \using \text{Lift}
    \end{prooftree}
    \qquad
    \begin{prooftree}
        \Gamma, f:\alpha\toF\gamma \vdash E:\delta
        \just \Gamma \vdash \Shift f\Colon E:\alpha_\gamma^\delta
        \using \text{Shift}
    \end{prooftree}
\\
    \begin{prooftree}
        \Gamma \vdash F:(\alpha\Fwd\beta_{\gamma3}^{\gamma2})_{\gamma1}^{\gamma0}
        \quad \Gamma \vdash E:\alpha_{\gamma2}^{\gamma1}
        \just \Gamma \vdash F\fwd E:\beta_{\gamma3}^{\gamma0}
        \using \Fwd{\text E}
    \end{prooftree}
    \qquad
    \begin{prooftree}
        \Gamma \vdash E:\alpha_{\gamma1}^{\gamma0}
        \quad \Gamma \vdash F:(\alpha\Bwd\beta_{\gamma3}^{\gamma2})_{\gamma2}^{\gamma1}
        \just \Gamma \vdash E\bwd F:\beta_{\gamma3}^{\gamma0}
        \using \Bwd{\text E}
    \end{prooftree}
\end{gather*}
\caption{A logical metalanguage with directionality and delimited control operators}
\label{fig:metalanguage}
\end{figure*}

\begin{figure}
\newcommand{\blank}[1][\enspace]{\langle#1\rangle}
\abovedisplayskip     =0pt
\abovedisplayshortskip=0pt
\begin{align*}
    \tag*{Values} V
        &::= U \mid \fun{x}E \mid \fun[\Delta]{x}E \\
    \tag*{Unknowns} U
        &::= c \mid UV \mid U\fwd V \mid V\bwd U \\
    \tag*{Contexts} C\blank
        &::= \blank
        \mid (C\blank)    E
        \mid  C\blank\fwd E
        \mid  C\blank\bwd F \\&\hphantom{{}::= \blank}
        \mid V    (C\blank)
        \mid V\fwd C\blank
        \mid V\bwd C\blank \\
    \tag*{Metacontexts} D\blank
        &::= \blank
        \mid C\blank[{[D\blank]}]
\end{align*}

\abovedisplayskip     =\smallskipamount
\abovedisplayshortskip=\smallskipamount
\belowdisplayskip     =0pt
\belowdisplayshortskip=0pt
\noindent Computations $E \evalsto E'$
\begin{align*}
    D\blank[{C\blank[\text{\ul{$(\fun      {x}E)     V$}}]}] &\enspace\evalsto\enspace D\blank[{C\blank[{E\{x\mapsto V\}}]}] \\
    D\blank[{C\blank[\text{\ul{$(\fun[\fwd]{x}E)\fwd V$}}]}] &\enspace\evalsto\enspace D\blank[{C\blank[{E\{x\mapsto V\}}]}] \\
    D\blank[{C\blank[\text{\ul{$V\bwd (\fun[\bwd]{x}E)$}}]}] &\enspace\evalsto\enspace D\blank[{C\blank[{E\{x\mapsto V\}}]}] \\
    D\blank[{C\blank[\text{\ul{$[V]                   $}}]}] &\enspace\evalsto\enspace D\blank[{C\blank[{V}]}] \\
    D\blank[{C\blank[\text{\ul{$\Shift f\Colon E      $}}]}] &\enspace\evalsto\enspace D\blank[{E\{f\mapsto\fun{x}[C\blank[x]]\}}]
\end{align*}
\caption{Reductions for the logical metalanguage}
\label{fig:reductions}
\end{figure}

Figure~\vref{fig:metalanguage} shows a logical metalanguage that formalizes the
basic ideas presented above.  It is in this language that denotations on this
page are written and reduced.  Refining Danvy and Filinski's original
shift-reset language, we distinguish between \emp{pure} and \emp{impure}
expressions.  An impure expression may incur control effects when evaluated,
whereas a pure expression only incurs control effects contained within reset
\citep{danvy-cps,danvy-transformation,nielsen-selective,thielecke-control}.
This distinction is reflected in the typing judgments: an impure judgment
\begin{equation}
    \Gamma\vdash E:\smash[t]{\alpha_\gamma^\delta}
\end{equation}
not only gives a type~$\alpha$ for~$E$ itself but also specifies two answer
types $\gamma$ and~$\delta$.  By contrast, a pure judgment
\begin{equation}
    \Gamma\vdash E:\alpha
\end{equation}
only gives a type~$\alpha$ for~$E$ itself.  As can be seen in the Lift rule,
pure expressions are polymorphic in the answer type.

As mentioned in~\S\ref{s:formalism}, the use of directionality in function types
to control word order is not new in linguistics, but the use of delimited
control operators to analyze quantification is.  It turns out that we can tie
the potential presence of control effects in function bodies to directionality.
That is, only directional functions---those whose types are decorated with tick
marks---are potentially impure; all non-directional functions we need to deal
with, including contexts captured by shift, are pure.  Another link between
directionality and control effects is that the $\Fwd$E and $\Bwd$E rules for
directional function application are not merely mirror images of each other: the
answer types $\gamma_0$ through~$\gamma_3$ are chained differently through the
premises.  This is due to left-to-right evaluation.

Having made the distinction between pure and impure expressions, we require in
our Shift rule that the body of a shift expression be pure.  This change from
Danvy and Filinski's original system simplifies the type system and the CPS
transform, but a shift expression $\Shift f\Colon E$ in their language may
need to be rewritten here to~$\Shift f\Colon[E]$.

The CPS transform for the metalanguage follows from the typing rules and is
standard; it supplies a denotational semantics.  The operational semantics for
the metalanguage specifies a computation relation between complete terms; it is
also standard and shown in Figure~\ref{fig:reductions}.

The present analysis is almost, but not quite, the direct-style analogue of
\citets{barker-continuations} CPS analysis.  Put in direct-style terms, Barker's
function bodies are always pure, whereas function bodies here can harbor control
effects.  In other words, function bodies are allowed to shift, as in the
determiner denotations in \eqref{e:every} and~\eqref{e:some}.  By contrast,
Barker uses \emph{choice functions} to assign meanings to determiners.

\section{Quantifier scope ambiguity}
\label{s:hierarchy}

Of course, natural language phenomena are never as simple as a couple of
programming language control operators.  Quantification is no exception, so to
speak.  For example, the sentence \phrase{Some student liked
every course} \eqref{e:sslec} is ambiguous between the following two readings.
\begin{gather}
\label{e:sslec-surface}\Exists{x} \Student(x) \land \Forall{y} \Course(y) \limplies x\bwd\Liked\fwd y\\
\label{e:sslec-inverse}\Forall{y} \Course(y) \limplies \Exists{x} \Student(x) \land x\bwd\Liked\fwd y
\end{gather}
In the \emp{surface scope} reading~\eqref{e:sslec-surface}, \phrase{some} takes
scope over \phrase{every}.  In the \emp{inverse scope}
reading~\eqref{e:sslec-inverse}, \phrase{every} takes scope over \phrase{some}.
Given that evaluation takes place from left to right, the shift
for \phrase{some student} is evaluated before the shift for \phrase{every
course}.  Our grammar thus predicts the surface scope reading but not the
inverse scope reading.  This prediction can be seen in the first few reductions
of the (unique) derivation for~\eqref{e:sslec}:
\begin{align}
    \rlap{$
        \bigl[ \left(\denote{\text{some}} \fwd \denote{\text{student}}\right)
          \bwd \denote{\text{liked}}
          \fwd \denote{\text{every}} \fwd \denote{\text{course}} \bigr]
        $}\enspace
\\\notag
    &= \bigl[ \bigl( \text{\ul{$
                (\fun[\fwd]{r} \Shift s\Colon \Exists{x} r(x) \land     s(x))
              \fwd
                \Student
              $}} \bigr)
\\\notag
    &\hspace*{3em}{}
         \bwd \Liked
         \fwd \bigl(
                (\fun[\fwd]{r} \Shift s\Colon \Forall{y} r(y) \limplies s(y))
              \fwd
                \Course
              \bigr)
       \bigr]
\\\notag
    &\evalsto
       \bigl[ \bigl( \text{\ul{$
                \Shift s\Colon \Exists{x} \Student(x) \land     s(x)
              $}} \bigr)
\\\notag
    &\hspace*{3em}{}
         \bwd \Liked
         \fwd \bigl(
                (\fun[\fwd]{r} \Shift s\Colon \Forall{y} r(y) \limplies s(y))
              \fwd
                \Course
              \bigr)
       \bigr]
\\\notag
    &\evalsto
       \bigl[ \Exists{x} \Student(x) \land
         \bigl(\fun{v} [v
\\\notag
    &\hspace*{3em}{}
         \bwd \Liked
         \fwd (
                (\fun[\fwd]{r} \Shift s\Colon \Forall{y} r(y) \limplies s(y))
              \fwd
                \Course
              )]\bigr)(x)
       \bigr]
    \hspace*{-.5em}
\end{align}
Regardless of what evaluation order we specify, as long
as our rules for semantic translation remain deterministic, they will only
generate one reading for the sentence.  Hence our theory fails to
predict the ambiguity of the sentence~\eqref{e:sslec}.

To better account for the data, we need to introduce some sort of nondeterminism
into our theory.  There are two natural ways to proceed.  First, we can allow
arbitrary evaluation order, not just left-to-right.  This change would render
our term calculus nonconfluent, a result unwelcome for most programming language
researchers but welcome for us in light of the ambiguous natural language
sentence~\eqref{e:sslec}.  This route has been pursued with some success by
\citet{barker-continuations} and \citet{de-groote-type}.  However, there are
empirical reasons to maintain left-to-right evaluation, one of which appears
in~\S\ref{s:polarity}.

\begin{figure*}[t]
\newcommand{\just}{\justifies}
\abovedisplayskip     =0pt
\abovedisplayshortskip=0pt
\begin{align*}
    \tag*{Directions} \Delta
        &::= \fwd \mid \bwd \\
    \tag*{Value types} \alpha,\beta,\gamma,\delta
        &::= \smash[t]{
             \Thing\{\vec{p}\} \mid \Bool\{\vec{p}\}
        \mid \alpha\toF\beta!n
        \mid \alpha\toD\beta!n} \\
    \tag*{Computation types} \alpha!0
        &::= \alpha, \qquad \alpha!(n + 1)
         ::= \smash{\alpha_{\gamma!n}^{\delta!n}} \\
    \tag*{Antecedents} \Gamma
        &::= x_1:\alpha_1,\dotsc,x_n:\alpha_n \\
    \tag*{Terms} E,F
        &::= c \mid x \mid \fun[n]{x}E \mid \fun[\Delta n]{x}E
        \mid FE \mid F\fwd E \mid E\bwd F \mid [E]^n \mid \Shift^n\!f\Colon E
\end{align*}

\abovedisplayskip     =\smallskipamount
\abovedisplayshortskip=\smallskipamount
\renewcommand{\just}{\justifies\strut}
\noindent Constants $c:\alpha$
\begin{gather*}
    \begin{prooftree}
        \just p : \Thing\{p\}
    \end{prooftree}
    \qquad
    \begin{prooftree}
        \just \forall p : \Bool\{p,\vec{q}\}\toF\Bool\{\vec{q}\}
    \end{prooftree}
    \qquad
    \begin{prooftree}
        \just \exists p : \Bool\{p,\vec{q}\}\toF\Bool\{\vec{q}\}
    \end{prooftree}
\\
    \begin{prooftree}
        \just \limplies : \Bool\{\vec{p}\}\toF\Bool\{\vec{q}\}\toF\Bool\{\vec{p}\cup\vec{q}\}
    \end{prooftree}
    \qquad
    \begin{prooftree}
        \just \land : \Bool\{\vec{p}\}\toF\Bool\{\vec{q}\}\toF\Bool\{\vec{p}\cup\vec{q}\}
    \end{prooftree}
\\
    \begin{prooftree}
        \just \Student : \Thing\{\vec{p}\}\toF\Bool\{\vec{p}\}
    \end{prooftree}
    \qquad
    \begin{prooftree}
        \just \Liked : \Thing\{\vec{p}\}\Fwd(\Thing\{\vec{q}\}\Bwd\Bool\{\vec{p}\cup\vec{q}\}{}_{\delta!m}^{\delta!m})_{\gamma!n}^{\gamma!n}
    \end{prooftree}
    \qquad
    \dotsm
\end{gather*}

\belowdisplayskip     =0pt
\belowdisplayshortskip=0pt
\renewcommand{\just}{\strut\justifies\strut}
\noindent Expressions $\Gamma \vdash E:\alpha!n$
\begin{gather*}
    \let\vcenter\vbox
    \begin{prooftree}
        c:\alpha
        \just \Gamma \vdash c:\alpha
        \using \text{Const}
    \end{prooftree}
    \qquad
    \begin{prooftree}
        \just \Gamma,x:\alpha \vdash x:\alpha
        \using \text{Var}
    \end{prooftree}
    \qquad
    \begin{prooftree}
        \Gamma \vdash E:\alpha_\alpha^\beta
        \just \Gamma \vdash [E]^0:\beta
        \using \text{Reset}
    \end{prooftree}
    \qquad
  \smash{
    \begin{prooftree}
        \Gamma \vdash E:\alpha_{\alpha_{\gamma!n}^{\gamma!n}}^{\beta!(n+1)}
        \just \Gamma \vdash [E]^{n+1}:\beta!(n+1)
        \using \text{Reset}
    \end{prooftree}
  }
\\
    \begin{prooftree}
        \Gamma, x:\alpha \vdash E:\beta!n
        \just \Gamma \vdash \fun[(\Delta)]{x}E:\alpha\toMaybeD\beta!n
        \using \toMaybeD\text{I}
    \end{prooftree}
    \qquad
    \begin{prooftree}
        \Gamma \vdash F:\alpha\toF\beta!n
        \quad \Gamma \vdash E:\alpha
        \just \Gamma \vdash FE:\beta!n
        \using \toF\text{E}
    \end{prooftree}
\\
    \let\vcenter\vtop
    \begin{prooftree}
        \Gamma \vdash E:\alpha
        \just \Gamma \vdash E:\alpha_\gamma^\gamma
        \using \text{Lift}
    \end{prooftree}
    \qquad
    \begin{prooftree}
        \Gamma \vdash E:\alpha_{\gamma1!n}^{\gamma0!n}
        \just \Gamma \vdash E:\alpha_{\beta_{\gamma2!n}^{\gamma1!n}}
                                    ^{\beta_{\gamma2!n}^{\gamma0!n}}
        \using \text{Lift}
    \end{prooftree}
    \qquad
    \begin{prooftree}
        \Gamma, f:\alpha\toF\gamma!n \vdash E:\delta!n
        \just \Gamma \vdash \Shift^n\!f\Colon E : \alpha_{\gamma!n}^{\delta!n}
        \using \text{Shift}
    \end{prooftree}
\\
    \begin{prooftree}
        \Gamma \vdash F:(\alpha\Fwd\beta_{\gamma3!n}^{\gamma2!n})_{\gamma1!n}^{\gamma0!n}
        \quad \Gamma \vdash E:\alpha_{\gamma2!n}^{\gamma1!n}
        \just \Gamma \vdash F\fwd E:\beta_{\gamma3!n}^{\gamma0!n}
        \using \Fwd{\text E}
    \end{prooftree}
    \qquad
    \begin{prooftree}
        \Gamma \vdash E:\alpha_{\gamma1!n}^{\gamma0!n}
        \quad \Gamma \vdash F:(\alpha\Bwd\beta_{\gamma3!n}^{\gamma2!n})_{\gamma2!n}^{\gamma1!n}
        \just \Gamma \vdash E\bwd F:\beta_{\gamma3!n}^{\gamma0!n}
        \using \Bwd{\text E}
    \end{prooftree}
\\
    \begin{prooftree}
        \Gamma \vdash F:(\alpha\Fwd\beta)
        \quad \Gamma \vdash E:\alpha
        \just \Gamma \vdash F\fwd E:\beta
        \using \Fwd{\text E}
    \end{prooftree}
    \qquad
    \begin{prooftree}
        \Gamma \vdash E:\alpha
        \quad \Gamma \vdash F:(\alpha\Bwd\beta)
        \just \Gamma \vdash E\bwd F:\beta
        \using \Bwd{\text E}
    \end{prooftree}
\end{gather*}
\caption{Extending the logical metalanguage to a hierarchy of control operators}
\label{fig:hierarchy}
\end{figure*}

\begin{figure*}
\newcommand{\blank}[1][\enspace]{\langle#1\rangle}
\abovedisplayskip     =0pt
\abovedisplayshortskip=0pt
\begin{align*}
    \tag*{Values} V
        &::= U \mid \fun{x}E \mid \fun[\Delta]{x}E \\
    \tag*{Unknowns} U
        &::= c \mid UV \mid U\fwd V \mid V\bwd U \\
    \tag*{Contexts} C\blank
        &::= \blank
        \mid (C\blank)    E
        \mid  C\blank\fwd E
        \mid  C\blank\bwd F
        \mid V    (C\blank)
        \mid V\fwd C\blank
        \mid V\bwd C\blank \\
    \tag*{Metacontexts at level~$n$} D^n\blank
        &::= \blank
        \mid D^{n+1}\blank[{D^{n+2}\blank[{\dotsm\blank[{C\blank[{[D^n\blank]^n}]}]\dotsm}]}]
\end{align*}

\abovedisplayskip     =\smallskipamount
\abovedisplayshortskip=\smallskipamount
\belowdisplayskip     =0pt
\belowdisplayshortskip=0pt
\noindent Computations $E \evalsto E'$
\begin{align*}
    D^0\blank[{D^1\blank[{\dotsm\blank[{C\blank[\text{\ul{$(\fun      {x}E)     V$}}]}]\dotsm}]}] &\enspace\evalsto\enspace D^0\blank[{D^1\blank[{\dotsm\blank[{C\blank[{E\{x\mapsto V\}}]}]\dotsm}]}] \\
    D^0\blank[{D^1\blank[{\dotsm\blank[{C\blank[\text{\ul{$(\fun[\fwd]{x}E)\fwd V$}}]}]\dotsm}]}] &\enspace\evalsto\enspace D^0\blank[{D^1\blank[{\dotsm\blank[{C\blank[{E\{x\mapsto V\}}]}]\dotsm}]}] \\
    D^0\blank[{D^1\blank[{\dotsm\blank[{C\blank[\text{\ul{$V\bwd (\fun[\bwd]{x}E)$}}]}]\dotsm}]}] &\enspace\evalsto\enspace D^0\blank[{D^1\blank[{\dotsm\blank[{C\blank[{E\{x\mapsto V\}}]}]\dotsm}]}] \\
    D^0\blank[{D^1\blank[{\dotsm\blank[{C\blank[\text{\ul{$[V]                   $}}]}]\dotsm}]}] &\enspace\evalsto\enspace D^0\blank[{D^1\blank[{\dotsm\blank[{C\blank[{V}]}]\dotsm}]}] \\
    D^0\blank[{D^1\blank[{\dotsm\blank[{D^n\blank[{D^{n+1}\blank[{\dotsm\blank[{C\blank[\text{\ul{$\Shift^n\!f\Colon E$}}]}]\dotsm}]}]}]\dotsm}]}] &\enspace\evalsto\enspace D^0\blank[{D^1\blank[{\dotsm\blank[{D^n\blank[{E\{f\mapsto\fun{x}[D^{n+1}\blank[{\dotsm\blank[{C\blank[x]}]\dotsm}]]^n\}}]}]\dotsm}]}]
\end{align*}
\caption{Reductions for the extended logical metalanguage}
\label{fig:reductions-hierarchy}
\end{figure*}

A second way to introduce nondeterminism is to maintain left-to-right evaluation
but generalize shift and reset to a \emp{hierarchy} of control operators
\citep{danvy-abstracting,barker-higher-order,shan-explaining}, leaving it
unspecified at which level on the hierarchy each quantificational phrase shifts.
Following \citeauthor{danvy-abstracting}, we extend our logical
metalanguage by superscripting every shift expression and pair of reset brackets
with a nonnegative integer to indicate a level on the control hierarchy.
Level~$0$ is the highest level (not the lowest).  When a shift expression at
level~$n$ is evaluated, it captures the current context of computation up to the
closest dynamically enclosing reset at level~$n$ or higher (smaller).  For
example, whereas the expression
\begin{equation}
    \bigl[3 \times [10 \times (\Shift^5\!f.\, 1 + f(2))]^5\bigr]^0
\end{equation}
evaluates to~$63$ as in~\eqref{e:reset-example}, the expression
\begin{equation}
    \bigl[3 \times [10 \times (\Shift^3\!f.\, 1 + f(2))]^5\bigr]^0
\end{equation}
evaluates to $1 + 3 \times 10 \times 2$, or~$61$.  The superscripts can be
thought of ``strength levels'' for shifts and resets.

\Citet{danvy-abstracting} give a denotational semantics for multiple levels of
delimited control using continuations of higher-order type.  We can take
advantage of that work in our quantificational
denotations~(\refrange{e:every}{e:some}) by letting them shift at any level.
The ambiguity of~\eqref{e:sslec} is then predicted as follows.  Suppose that
\phrase{some student} shifts at level~$m$ and \phrase{every course} shifts at
level~$n$.
\example{Some$^m$ student liked every$^n$ course.}
If $m\le n$, the surface scope reading~\eqref{e:sslec-surface} results.  If
$m>n$, the inverse scope reading~\eqref{e:sslec-inverse} results.  In general,
a quantifier that shifts at a higher level always scopes over another that
shifts on a lower level, regardless of which one is evaluated first.  This
way, evaluation order does not determine scoping possibilities among quantifiers
in a sentence unless two quantifiers happen to shift at the same level.

To summarize the discussion so far, whether we introduce nondeterministic
evaluation order or a hierarchy of delimited control operators, we can account
for the ambiguity of the sentence~\eqref{e:sslec}, as well as more complicated
cases of quantification in English and Mandarin \citep{shan-quantifier}.  For
example, both the nondeterministic evaluation order approach and the control
hierarchy approach predict correctly that the sentence below, with three
quantifiers, is $5$-way ambiguous.
\example{\label{e:three-quantifiers}Every representative of a company saw most samples.}
Despite the fact that there are three quantifiers in this sentence and $3! = 6$,
this sentence has only $5$ readings.  Because \phrase{a company} occurs within
the restrictor of \phrase{every representative of a company}, it is incoherent
for \phrase{every} to scope over \phrase{most} and \phrase{most}
over~\phrase{a}.  The reason neither approach generates such a reading can
be seen in the denotation of \phrase{every} in~\eqref{e:every}: ``$\forall x$''
is located immediately above ``$\limplies$'' in the abstract syntax, with no
intervening control delimiter, so no control operator can insert any material
(such as \phrase{most}\hyp quantification over samples) in between.

There exist in the computational linguistics literature algorithms for computing
the possible quantifier scopings of a sentence like~\eqref{e:three-quantifiers}
\citep[][followed by
\citealp{moran-quantifier,lewin-quantifier}]{hobbs-algorithm}.  Having related
quantifier scoping to control operators, we gain a denotational
understanding of these algorithms that accords with our theoretical intuitions
and empirical observations.

An extended logical metalanguage with an infinite hierarchy of control operators
is shown in Figure~\ref{fig:hierarchy}.  This system is more complex than the
one in Figure~\ref{fig:metalanguage} in two ways.  First, instead of making a
binary distinction between pure and impure expressions, we use a number to
measure ``how pure'' each expression is.  An expression is pure up to level~$n$
if it only incurs control effects at levels above~$n$ when evaluated.  Pure
expressions are the special case when $n=0$.  The purity level of an expression
is reflected in its typing judgment: a judgment
\begin{equation}
    \Gamma\vdash E:\alpha!n
\end{equation}
states that the expression~$E$ is pure up to level~$n$.  Here $\alpha!n$ is
a \emph{computation type} with $n$ levels: as defined in the figure, it
consists of $2^{n+1}-1$ value types that together specify how a computation
that is pure up to level~$n$ affects answer types between levels~$0$
and~$n-1$.  In the special case where $n=1$, the computation type $\alpha!1$ is
of the familiar form~$\alpha_\gamma^\delta$.

In the previous system in Figure~\ref{fig:metalanguage}, directional functions
are always impure (that is, pure up to level~$1$) while non-directional
functions are always pure (that is, pure up to level~$0$).  In the current
system, both kinds of functions declare in their types up to what level their
bodies are pure.  For example, the determiners \phrase{every} and \phrase{some},
now allowed to shift at any level, both have not just the type
\begin{gather}
    \label{e:det-type-pure-restrictor}
    (\Thing \toF \Bool) \Fwd
    \Thing     _{\Bool_{\gamma!n}^{\delta!n}}
               ^{\Bool_{\gamma!n}^{\delta!n}},
\end{gather}
but also the type
\begin{equation}
    \label{e:det-type-impure-restrictor}
    (\Thing \toF \Bool_{\gamma1!n}^{\gamma0!n}) \Fwd
    \Thing     _{\Bool_{\gamma2!n}^{\gamma1!n}}
               ^{\Bool_{\gamma2!n}^{\gamma0!n}}
\end{equation}
(but see the second technical complication below).  As the argument type
$\Thing \toF \smash{\Bool_{\gamma1!n}^{\gamma0!n}}$ above shows,
the first argument to these determiners, the restrictor, is non-directional
yet can be impure (that is, pure up to level $n+1$).

To traverse the control hierarchy, we add a new Reset rule, which makes an
expression more pure, and a new Lift rule, which makes an expression less pure.
(Consecutive nested resets like $\bigl[[E]^{n+1}\bigr]^n$ can be abbreviated to
$[E]^n$ without loss of coherence.)

A second complication in this system, in contrast to
Figure~\ref{fig:metalanguage}, is that we can no longer encode logical
quantification using a higher-order constant like
$\forall:(\Thing\toF\Bool)\toF\Bool$, because such a constant requires its
argument---the logical formula to be quantified---to be a pure function.  This
requirement is problematic because it is exactly the impurity of quantified
logical formulas that underlies this account of quantifier scope ambiguity.  On
one hand, we want to quantify logical formulas that are impure; on the other
hand, we want to rule out expressions like
\begin{equation}
    \label{e:leak}
    \Forall{x} \Shift f\Colon x,
\end{equation}
where the logical variable~$x$ ``leaks'' illicitly into the surrounding context.
This issue is precisely the problem of classifying open and closed terms in
staged programming (see \citealp{taha-environment} and references therein): the
types $\Thing$ and $\Bool$ really represent not individuals or truth values but
staged programs that compute individuals and truth values.  For this paper, we
adopt the simplistic solution of adjoining to these types a set of free logical
variables for tracking purposes, denoted~$p,q,\dotsc$.  Unfortunately, we also
need to stipulate that these logical variables be freshly $\alpha$-renamed
(``created by gensym'') for each occurrence of a quantifier in a sentence.

As before, the CPS transform and reductions for this metalanguage are
standard; the latter appears in Figure~\ref{fig:reductions-hierarchy}.

The present analysis is almost, but not quite, the direct-style analogue of
\citets{shan-explaining} CPS analysis, even though both use a control
hierarchy.  Each level in \citeauthor{shan-explaining}'s hierarchy is
intuitively a staged computation produced at one level higher.  More
concretely, a computation type with $n$ levels in that system has the shape
\begin{equation}
    (\alpha_{\gamma1}^{\gamma0})_{\delta1}^{\delta0}
\end{equation}
rather than the shape
\begin{equation}
    \alpha_{\gamma1_{\delta3}^{\delta2}}
          ^{\gamma0_{\delta1}^{\delta0}}.
\end{equation}
The issue above of how to encode logical quantification over impure formulas
receives a more satisfactory treatment in Shan and Barker's system: no
stipulation of $\alpha$-renaming is necessary, because there is no analogue
of~\eqref{e:leak} to prohibit.  The relation between that system and staged
programming with effects has yet to be explored.

\section{Polarity sensitivity}
\label{s:polarity}

Because the analysis so far focuses on the truth\hyp conditional meaning of
quantifiers, it equates the determiners \phrase{a} and
\phrase{some}---both are existential quantifiers with the type and
denotation in~\eqref{e:some}.  Furthermore, sentences like \phrase{Has anyone
arrived?} suggest that the determiner \phrase{any} also means the same thing as
\phrase{a} and \phrase{some}.  To the contrary, though, the determiners
\phrase{a}, \phrase{some}, and \phrase{any} are not always interchangeable in
their existential usage.  The sentences and readings in~\eqref{e:polarity} show
that they take scope differently relative to negation (in these cases the
quantifier~\phrase{no}).
\examples[l@{}l@{}X]{
\ex\label{e:polarity}
&\ey\label{e:polarity-unordered-first}\label{e:no-some}
    &No student liked some course.\hfill (unambiguous $\exists\neg$)\\
&\ey\label{e:no-a}
    &No student liked a course.\hfill (ambiguous $\neg\exists$, $\exists\neg$)\\
&\ey\label{e:no-any}
    &No student liked any course.\hfill (unambiguous $\neg\exists$)\\
&\ey\label{e:some-no}
    &Some student liked no course.\hfill (unambiguous $\exists\neg$)\\
&\ey\label{e:polarity-unordered-last}\label{e:a-no}
    &A student liked no course.\hfill (ambiguous $\neg\exists$, $\exists\neg$)\\
&\ey\label{e:any-no}
    &\<*Any student liked no course.\hfill (unacceptable)}

The determiner \phrase{any} is a \emph{negative polarity item}: to a first
approximation, it can occur only in \emph{downward\hyp
entailing} contexts, such as under the scope of a \emph{monotonically
decreasing} quantifier \citep{ladusaw-polarity}.
A quantifier~$q$, of type $(\Thing\toF\Bool)\toF\Bool$, is
monotonically decreasing just in case
\begin{equation}
    \Forall{s_1} \Forall{s_2}
    \bigl(\Forall{x} s_2(x) \limplies s_1(x)\bigr)
        \limplies q(s_1) \limplies q(s_2).
\end{equation}
The quantificational noun phrases \phrase{no student} and \phrase{no course} are
monotonically decreasing since, for instance, if no student liked any course in
general, then no student liked any computer science course in particular.

Whereas \phrase{any} is a negative polarity item, \phrase{some} is
a \emph{positive polarity item}. Roughly speaking, \phrase{some} is allergic to
downward\hyp entailing contexts (especially those with an overtly negative word
like~\phrase{no}).  These generalizations regarding polarity items cover the
data in~(\refrange{e:polarity-unordered-first}{e:polarity-unordered-last}): in
principle, goes the theory, all these sentences are ambiguous between two
scopings, but the polarity sensitivity of \phrase{some} and \phrase{any} rule
out one scoping each in \eqref{e:no-some}, \eqref{e:no-any}, \eqref{e:some-no},
and~\eqref{e:any-no}.  These four sentences are thus predicted to be
unambiguous, but it remains unclear why \eqref{e:any-no} is downright
unacceptable.

In the type-theoretic tradition of linguistics, polarity sensitivity is
typically implemented by splitting the answer type $\Bool$ into several types,
each a different functor applied to $\Bool$, that are related by subtyping
\citep{fry-proof,bernardi-reasoning,bernardi-generalized}.  For instance, to
differentiate the determiners in~\eqref{e:polarity} from each other in our
formalism, we can add the types $\BoolPos$ and $\BoolNeg$ alongside
$\Bool$, such that both are supertypes of $\Bool$ (but not of each other).
\begin{equation}
\label{e:subtype}
    \begin{prooftree} \justifies \Bool \le \BoolPos \end{prooftree}
    \qquad
    \begin{prooftree} \justifies \Bool \le \BoolNeg \end{prooftree}
\end{equation}
We also extend the subtyping relation between (value and computation) types with
the usual closure rules,
and allow implicit coercion from a subtype to a supertype.
\begin{gather}
    \begin{prooftree}
        \strut\justifies\strut \alpha \le \alpha
    \end{prooftree}
    \qquad
    \begin{prooftree}
        \alpha' \le \alpha
        \quad \beta!n \le \beta'!n
        \strut\justifies\strut \alpha\toMaybeD\beta!n \le \alpha'\toMaybeD\beta'!n
    \end{prooftree}
\displaybreak[0]
\\
    \begin{prooftree}
        \alpha' \le \alpha
        \quad \gamma'!n \le \gamma!n
        \quad \delta!n \le \delta'!n
        \justifies \alpha_{\gamma!n}^{\delta!n}
            \le \alpha'{}_{\gamma'!n}^{\delta'!n}
    \end{prooftree}
\displaybreak[1]
\\
    \begin{prooftree}
        \Gamma \vdash E:\alpha!n
        \quad \alpha!n\le\beta!n
        \strut\justifies\strut \Gamma \vdash E:\beta!n
        \using \text{Sub}
    \end{prooftree}
\end{gather}
We then add a side condition to Reset, requiring that the
produced answer type be $\Bool$ or $\BoolPos$, not $\BoolNeg$.
\begin{gather}
    \begin{prooftree}
        \Gamma \vdash E : \alpha_\alpha^\beta
        \justifies \Gamma \vdash [E] : \beta
        \using \text{Reset}
        \qquad \text{where $\beta \le \BoolPos$}
    \end{prooftree}
\displaybreak[0]
\\
    \begin{prooftree}
        \Gamma \vdash E : \alpha_{\alpha_{\gamma!n}^{\gamma!n}}^{\beta!(n+1)}
        \justifies \Gamma \vdash [E] : \beta!(n+1)
        \using \text{Reset}
        \qquad \text{where $\beta \le \BoolPos$}
    \end{prooftree}
\end{gather}
Finally, we refine the types of our determiners
from~\eqref{e:det-type-pure-restrictor} to
\begingroup
\newcommand{\transition}[3]{%
    \denote{\text{#1}}
    &: (\Thing\toF\Bool)\Fwd\Thing_{\smash[b]{#2}_{\gamma!n}^{\delta!n}}
                                  ^{\smash[b]{#3}_{\gamma!n}^{\delta!n}}}
\begin{align}
\label{e:poldet-first}
\transition{no}{\BoolNeg}{\Bool}, \displaybreak[0] \\
\transition{some}{\BoolPos}{\BoolPos}, \displaybreak[1] \\
\transition{a}{\Bool}{\Bool}, \displaybreak[0] \\
\transition{any}{\BoolNeg}{\BoolNeg}.
\label{e:poldet-last}
\end{align}
\endgroup

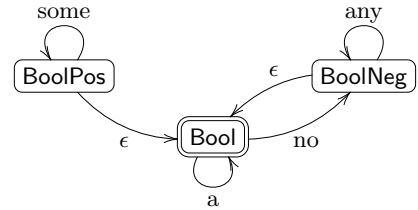
\begin{figure}[htbp]
\let\labelstyle\objectstyle
\centering
$
  \xymatrix @R=1em{
    *+[F-:<\jot>]{\BoolPos} \ar@/_1em/[dr]_{\epsilon} \ar@(ur,ul)[]_{\text{some}}
    &&
    *+[F-:<\jot>]{\smash[b]{\BoolNeg}} \ar@/_1em/[dl]_{\epsilon} \ar@(ur,ul)[]_{\text{\smash[b]{any}}}
    \\
    &
    *+[F-:<\jot>]+<\jot,\jot>[F-:<1.5\jot>]{\Bool} \ar@/_1em/[ur]_{\text{no}} \ar@(dl,dr)[]_{\text{a}}
  }
$
\caption{An automaton of answer-type transitions}
\label{fig:automaton}
\end{figure}

The chain of answer-type transitions from one quantificational expression to the
next acts as a finite-state automaton, shown in Figure~\vref{fig:automaton}.
The states of the automaton are the three supertypes of~$\Bool$; the
$\epsilon$\hyp transitions are the two subtyping relations in~\eqref{e:subtype};
and the non-$\epsilon$ transitions are the determiners
in~(\refrange{e:poldet-first}{e:poldet-last}).

This three-state machine enforces polarity constraints as follows.  Any valid
derivation for a sentence assigns each of its quantifiers to shift at
a certain level in the control hierarchy.  For each level, the quantifiers at
that level---in the order in which they are evaluated---must form
\begin{itemize*}
    \item either a path from $\BoolPos$ to $\Bool$ in the state machine, in
          other words a string of determiners matching the regular expression
          ``$\text{some}^*\,(\text{a}\,|\,\text{no}\;\text{any}^*)^*$'';
    \item or a path from $\Bool$ to $\Bool$ in the state machine, in other
          words a string of determiners matching the regular expression
          ``$(\text{a}\,|\,\text{no}\;\text{any}^*)^*$''.
\end{itemize*}
Furthermore, the $\BoolPos$-to-$\Bool$ levels in the hierarchy must all be
higher than the $\Bool$-to-$\Bool$ levels.  Every assignment of quantifiers to
levels that satisfies these conditions gives a reading for the sentence, in
which quantifiers at higher levels scope wider, and, among quantifiers at the
same level, ones evaluated earlier scope wider.

Consider now the two alternative ways to characterize scope ambiguity suggested
in \S\ref{s:hierarchy}.  The first approach is to allow arbitrary evaluation
order (and use a degenerate control hierarchy of one level only).  If we take
this route, we can account for all of the acceptability and ambiguity judgments
in~(\refrange{e:polarity-unordered-first}{e:polarity-unordered-last}), but we
cannot distinguish the acceptable sentence \eqref{e:no-any} from the
unacceptable \eqref{e:any-no}.  In other words, it would be a mystery how the
acceptability of a sentence hinges on the linear order in which the quantifiers
\phrase{no} and \phrase{any} appear.  This mystery has been noted by
\citet[\S9.2]{ladusaw-polarity} and \citet[\S8.2]{fry-proof} as a defect in
current accounts of polarity sensitivity.

The second approach, using a control hierarchy with multiple levels, fares
better by comparison.%
\footnote{Although this paper uses Danvy and Filinski's control hierarchy,
    polarity sensitivity can be expressed equally well in Shan and Barker's
    system.}
We can stick to left-to-right evaluation, under which---as
desired---an \phrase{any} must be preceded by a \phrase{no} that scopes over it
with no intervening \phrase{a} or~\phrase{some}.  Indeed, the variations in
ambiguity and acceptability among sentences in~\eqref{e:polarity} are completely
captured.  For intuition, we can imagine that the hearer of a sentence must
first process the trigger for a downward\hyp entailing context, like
\phrase{no}, before it makes sense to process a negative polarity item, like
\phrase{any}.%
\footnote{The syntactic distinction among the types $\Bool$, $\BoolPos$, and
    $\BoolNeg$ may even be semantically interpretable via the formulas-as-types
    correspondence, but the potential for such a connection has only been
    briefly explored \citep{bernardi-polarity} and we do not examine it here.
    In this connection, \citet{krifka-semantics} and others have proposed on
    pragmatic grounds that determiners like \phrase{any} are negative polarity
    items because they indicate extreme points on a scale.}
Intuition aside, the programming\hyp language notion of evaluation order
provides the syntactic hacker of formal types with a new tool with which to
capture observed regularities in natural language.

\section{Linguistic side effects}
\label{s:beyond}

This paper outlines how quantification and polarity sensitivity in natural
language can be modeled using delimited continuations.  These two examples
support my claim that the formal theory and computational intuition we have for
continuations can help us construct, understand, and maintain linguistic
theories.  To be sure, this work is far from the first time insights from
programming languages are applied to natural language:
\begin{itemize*}
    \item It has long been noted that the intensional logic in which Montague
          grammar is couched can be understood computationally
          \citep{hobbs-making,hung-semantics}.
    \item \emp{Dynamic semantics} \citep{groenendijk-dynamic}, which relates
          anaphora and discourse in natural languages to nondeterminism and
          mutable state in programming languages \citep{van-eijck-programming},
          has been applied to a variety of natural language phenomena, such as
          verb-phrase ellipsis
          \citep{van-eijck-verb-phrase,gardent-dynamic,hardt-dynamic}.
\end{itemize*}
However, the link between natural language and continuations has only
recently been made explicit, and this paper's use of control operators for
a direct-style analysis is novel.

The analyses presented here are part of a larger project, that of
relating computational side effects to \emp{linguistic side effects}.  The term
``computational side effect'' here covers all programming language features
where either it is unclear what a denotational semantics should look like, or
the ``obvious'' denotational semantics (such as making each arithmetic
expression denote a number) turns out to break referential transparency.
A computational side effect of the first kind is jumps to labels; one of the
second kind is mutable state.  By analogy, I use the term ``linguistic side
effects'' to refer to aspects of natural language where either it is unclear
what a denotational semantics should look like, or the ``obvious'' denotational
semantics (such as making each clause denote whether it is true) turns out to
break referential transparency.  Besides quantification and polarity
sensitivity, some examples are:
\examples[l@{}l@{}X@{}>(r<)]{
\ex\label{e:nl}
&\ey \label{e:nl-intensionality}
     &Bob \emph{thinks} Alice likes CS187. &Intensionality \\
&\ey &\emph{A man} walks. \emph{He} whistles.  &Variable binding \\
&\ey \label{e:nl-interrogatives}
     &\emph{Which} star did Alice see?  &Interrogatives \\
&\ey &Alice \emph{only} saw \emph{\textsc{Venus}}.  &Focus \\
&\ey &\emph{The king of France} whistles. &Presuppositions}

To study linguistic side effects, I propose to draw an analogy between them and
computational side effects.  Just as computer scientists want to express all
computational side effects in a uniform and modular framework and study how
control interacts with mutable state \citep{felleisen-revised}, linguists want
to investigate properties common to all linguistic side effects and study
how quantification interacts with variable binding.  Furthermore, just as
computer scientists want to relate operational notions like evaluation order
and parameter passing to denotational models like continuations and monads,
linguists want to relate the dynamics of information in language processing to
the static definition of a language as a generative device.  Whether this
analogy yields a linguistic theory that is empirically adequate is an open
scientific question that I find attractive to pursue.

\section{Acknowledgments}

\noindent
Thanks to Stuart Shieber, Chris Barker, Raffaella Bernardi, Barbara Grosz,
Pauline Jacobson, Aravind Joshi, William Ladusaw, Fernando Pereira, Avi Pfeffer,
Chris Potts, Norman Ramsey, Dylan Thurston, Yoad Winter, and anonymous
referees.  This work is supported by the United States National Science
Foundation Grants IRI-9712068 and BCS-0236592.

\bibliographystyle{mcbride}
\bibliography{ccshan}
\setlength{\dimen0}{\depthof{g}}
\vspace*{-\dimen0}

\end{document}